# Text to Artistic Image Generation

**Qinghe Tian[1], Pr. Jean-Claude Franchitti[2]**
[1]University of Waterloo
[2]New York University Courant Institute
teresa.tian@uwaterloo.ca, jcf@cs.nyu.edu

## Abstract

Painting is one of the ways for people to express their ideas, but what if people with disabilities in hands want to paint? To tackle this challenge, we create an end-to-end solution that can generate artistic images from text descriptions. However, due to the lack of datasets with paired text description and artistic images, it is hard to directly train an algorithm which can create art based on text input. To address this issue, we split our task into three steps: (1) Generate a realistic image from a text description by using Dynamic Memory Generative Adversarial Network (M. Zhu et al. 2019), (2) Classify the image as a genre that exists in the WikiArt dataset ("WikiArt.Org - Visual Art Encyclopedia" n.d.) using Resnet (He et al. 2015), (3) Select a style that is compatible with the genre and transfer it to the generated image by using neural artistic stylization network (Ghiasi et al. 2017). The code associated with this project can be found at Github.

## 1 Introduction

Generating realistic images from text descriptions is a trendy research field in computer vision. A variety of network architectures have been proposed to address this topic. However, few of them focus on generating artistic images based on the text input. Therefore, we built an application that produces photo-quality images and subsequently transfers styles to them.

To be more specific, our algorithm first takes a sentence that describes a specific scene and feeds it into a Dynamic Memory GAN that generates images with realistic details. Then, based on the content of the generated image, we recommend to users several compatible artistic styles and allow users to choose one among them. A painting that uses that style is then selected randomly and passed to the style transfer network which outputs the desired stylized image. The style transfer steps may be repeated multiple times to apply additional styles to the generated image.

## 2 Related Works

### 2.1 Text to realistic image generation

With the recent success of GANs (Goodfellow et al. 2014), a variety of methods have been proposed to handle image generation. Within this field, generating images based on text descriptions has gained interest in the research community.

Most of the traditional GANs architectures encode the text embedding into a global sentence vector, which is later used as a condition in a conditional GAN (Reed et al. 2016; Zhang et al. 2016; 2018). Although these methods result in impressive results, they face limitations when generating images with complex sceneries (e.g., images in the COCO dataset). Different approaches have been proposed to address this issue. AttnGAN (Xu et al. 2017) uses an attention mechanism to focus on specific words when drawing each sub-region. StackGAN (Zhang et al. 2016) and StackGAN++ (Zhang et al. 2018) generate images in two stages – a low-resolution image is generated initially, and a high-resolution image is obtained by refining the initial image. Other approaches rely on the semantic layout (bounding boxes and object shapes) constructed from text (W. Li et al. 2019), or introduce a bidirectional translation to perform text-to-image generation and image captioning (Sah et al. 2018).

Recently, the attention mechanism and multi-stage methods have gained popularity. However, there are two main problems with these approaches (M. Zhu et al. 2019). First, the generation of a high-resolution image is heavily dependent on the previously generated low-resolution image. Second, the same word representations are employed as part of the various image refinement processes. To cope with low-quality images generated at the initial stage and the sometimes-conflicting dynamic selection of most relevant words, DM-GAN (M. Zhu et al. 2019) utilizes a key-value memory structure (Miller et al. 2016) with a GAN framework and a memory writing gate to resolve both of the issues mentioned above.

### 2.2 Image Classification

Deep learning models have been used for a long time in the field of image recognition. Among them, Convolutional Neural Networks have become the leading architecture (Rawat and Wang 2017). Fueled by GPUs and large datasets,

CNN has achieved significant breakthroughs (He et al. 2015; He and Sun, n.d.). The mainstream approach to improve the accuracy consists of extending the network in depth and size, while using a dropout mechanism to prevent overfitting (Simonyan and Zisserman 2015; Szegedy et al. 2014).

However, despite high memory usage, degradation hinders the construction of a suitably deep network. Surprising, the degradation is not caused by overfitting; rather, as the depth of the network increases, the volume of training and testing errors increases (He et al. 2015; He and Sun, n.d.). Introducing a deep *residual learning* framework appears to address this issue. The framework utilizes residual blocks to create shortcut connections in the flow of information and implements residual mapping, which facilitates network optimization (He et al. 2015; Targ, Almeida, and Lyman 2016).

### 2.3 Style Transfer

Transferring the artistic style of a particular image to another has also been heavily researched in recent years. It has been proven that a trained image classifier can be applied to enable style transfer. For example, a VGG-19 optimized for object recognition is used in [21] to extract features from textured images and generate new textures. Later, (Gatys, Ecker, and Bethge 2015a) extends this idea to style transfer by adding a constraint that pertains to preserving the content of images.

Optimizing the resulting algorithm is computationally expensive. Several approaches address this problem by constructing a *feedforward style transfer network* (Johnson, Alahi, and Fei-Fei 2016; C. Li and Wand 2016; Ulyanov, Vedaldi, and Lempitsky 2017). However, by doing so, flexibility is significantly reduced, as the style transfer network is limited to a single style and separate style transfer networks have to be built for each style trained on.

Moreover, since each style is learnt separately, these methods fail to study shared features across all paintings. To increase the training speed without using a secondary network, (Ghiasi et al. 2017) introduces a new algorithm that can operate in real-time on unseen paintings while generalizing the style transferability to include unobserved styles.

## 3 Solution Approach

### 3.1 Text to image generation

**DAMSM Encoder**
The first step is to connect the content of images with corresponding text descriptions correctly. A widely applied approach is to encode natural language concepts using a Deep Attentional Multimodal Similarity Model (DAMSM) (Xu et al. 2017).

The text encoder, a bi-directional Long Short-Term Memory (Schuster and Paliwal 1997), extracts semantic vectors from text descriptions and creates a feature vector for each word and a global sentence vector. The image encoder, a Convolutional Neural Network, extracts global features and regional features of images. The attention model computes the relevance between each word in the sentence and sub-regions of images and the overall mapping of the description to the entire image. The DAMSM learns the attention model in a manner that only supervises the overall but not the regional matching.

**Conditioning Augmentation**
Conditioning augmentation (Zhang et al. 2016) is a technique designed to deal with the problem of overfitting by yielding more diverse samples from the dataset. Additional conditioning variables are sampled from an independent Gaussian distribution $\mathcal{N}(\mu(s), \Sigma(s))$, where s represents the text embedding. We also utilize Kullback-Leibler divergence between the standard Gaussian distribution and the Gaussian distribution of training data to preserve smoothness further. Hence, the loss of conditioning augmentation can be defined as:

$$L_{CA} = D_{KL}\left(\mathcal{N}(\mu(s), \Sigma(s)) \parallel \mathcal{N}(0, I)\right). \quad (1)$$

**DM-GAN**

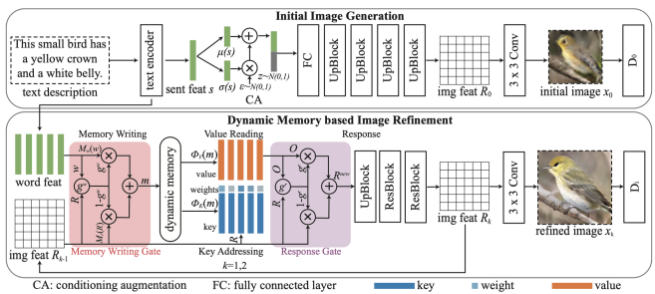

Figure 1. The architecture of DM-GAN (M. Zhu et al. 2019)

As shown in the Figure 1, the network has two stages: *initial image generation* and *dynamic memory-based image refinement*. During the first stage, the text encoder transforms text into internal representations, including a sentence feature $s$ and a few words features $W$, and a random noise vector $z$. The sentence feature is then used by the generator to predict an image feature $R_0$: $R_0 = G_0(z, s)$ which is corresponding to the rough shape of the initial image $x_0$. The second stage takes the image feature and word features resulting from the first stage to generate more realistic images $x_i$: $x_i = G_i(R_{i-1}, W)$.

Before receiving the final output, four key steps (memory writing, key addressing, value reading, and response) in the second stage refine images. At the memory writing stage, word features are encoded into dynamic memory. To select the most relevant words, the memory writing gate computes the significance of each word by combining image features $R_i$ with word features $W$. The memory writing gate subsequently writes the significance of each word into a memory slot. The key addressing stage accesses memory slots by using key memory and calculates the similarity probability between each memory slot and its corresponding image feature. The value reading stage outputs the memory representation by summing up memory features based on the similarity probability. The response stage receives the output and combines it with the current image feature to produce a new image feature, which is then upscaled into a high-resolution image by the up-sampling layer.

## 3.2 Image Classification

The second step in our approach is to choose a painting style for the image generated by DM-GAN. We assume that artwork with similar content as the realistic image will be suitable to act as the style image. For example, a photo of landscape fits perfectly for styles like impressionism. Therefore, we classify generated image based on the genre labels (i.e., groupings by objects and themes depicted by paintings) provided in the *WikiArt* dataset ("Artworks by Genre - WikiArt.Org" n.d.), and then select the most popular styles that are compatible with the genre.

To complete this step, we use the ResNet architecture (He et al. 2015). Instead of directly learning the mapping $H(x)$ among several staked layers, ResNet introduces the concept of residual learning: approximating a residual function $F(x) = H(x) - x$. This approach allows a deeper network to be sufficiently optimized. Although we aim at training a network that can categorize realistic images into genres that exist in the *WikiArt*, we lack an actual dataset. However, there must be shared patterns among images with similar content, even if they have distinct styles. Therefore, we train the network on paintings in the *WikiArt* dataset despite the fact that our final inputs are realistic photos.

## 3.3 Style Transfer

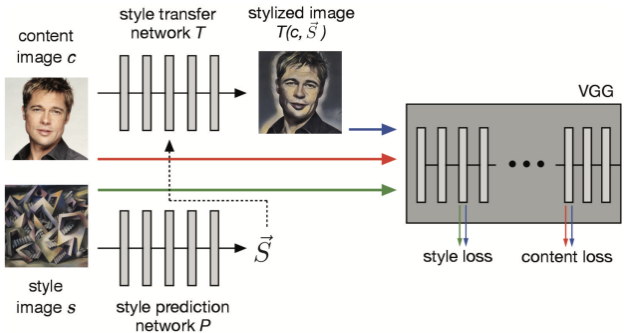

Figure 2. The architecture of style transfer network (Ghiasi et al. 2017)

When transferring the style of a painting to a photograph or realistic image, we generate a new image $x$ based on the content image $c$ and the style image $s$. The first challenge is to define the representations of content and style. According to (Gatys, Ecker, and Bethge 2015a), the content and style of an image are defined as high-level features and low-level features respectively extracted by an image recognition network. The more similar images are in content, the closer the high-level features are in terms of Euclidean distance. The more similar images are in style, the closer the low-level features in Gram matrix expressing are. Therefore, the content loss and style loss for the *style transfer network* are denoted as:

$$L_c(x,c) = \sum_{j \in C} \frac{1}{n_i} \| f_j(x) - f_j(c) \|_2^2, \quad (2)$$

$$L_s(x,s) = \sum_{i \in S} \frac{1}{n_i} \| \mathcal{G}[f_i(x)] - \mathcal{G}[f_i(s)] \|_F^2, \quad (3)$$

where $f_l(x)$ are the network activations at layer $l$, $n_l$ is the total number of units at layer $l$, and $G[f_l(x)]$ is the Gram matrix associated with the layer $l$ activations.

A traditional approach to increase the accuracy, other than training a new network for each artistic style, is to learn normalization parameters that are specific to different styles (Dumoulin, Shlens, and Kudlur 2017). However, this method only allows networks to work on explicitly trained styles. To generalize the style transferability to include unseen painting styles, we construct a *style prediction network* (Figure 2), composed of an Inception-v3 architecture (Szegedy et al. 2016) with two fully connected layers. This network can predict a normalization vector $\vec{S}$ that is specific to the input image $s$.

## 4 Experiments and Results

For the text to image generation stage, we directly applied the pre-trained model provided by DM-GAN ("MinfengZhu/DM-GAN" n.d.). As shown in Table 1, the model achieves 32.43 IS, 92.23% R-precision, and 24.24 FID on COCO 2014 dataset. Figure 3 shows several examples of text-to-image synthesis.

| Dataset | IS | R-precision | FID |
|---|---|---|---|
| CUB | 4.71 ± 0.06 | 76.58% ± 0.53% | 11.91 |
| COCO | 32.43 ± 0.58 | 92.23% ± 0.37% | 24.24 |

Table 1. Performance of Pretrained DM-GAN
IS = Inception Score, FID = Fréchet Inception Distance.

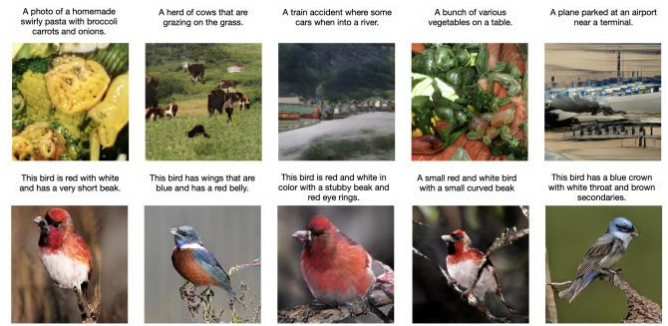

Figure 3. Images generated by pretrained DM-GAN.

For the image classification, we used a ResNet-18 pre-trained on ImageNet and fine-tuned its weights based on paintings from 10 popular genres in the *WikiArt* dataset. After training for 25 epochs, we achieved top accuracy of 75.92%. Although the accuracy can be further improved by using a deeper ResNet architecture or via training for more epochs, a perfect classifier on *WikiArt* dataset may not achieve an ideal result when classifying realistic photos. Therefore, to prevent overfitting, we merely train the network for a few epochs. Samples of classification result are shown in Figure 4.

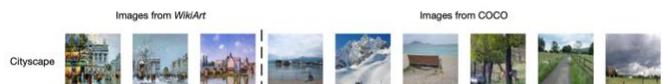

Figure 4. Classification result of ResNet.

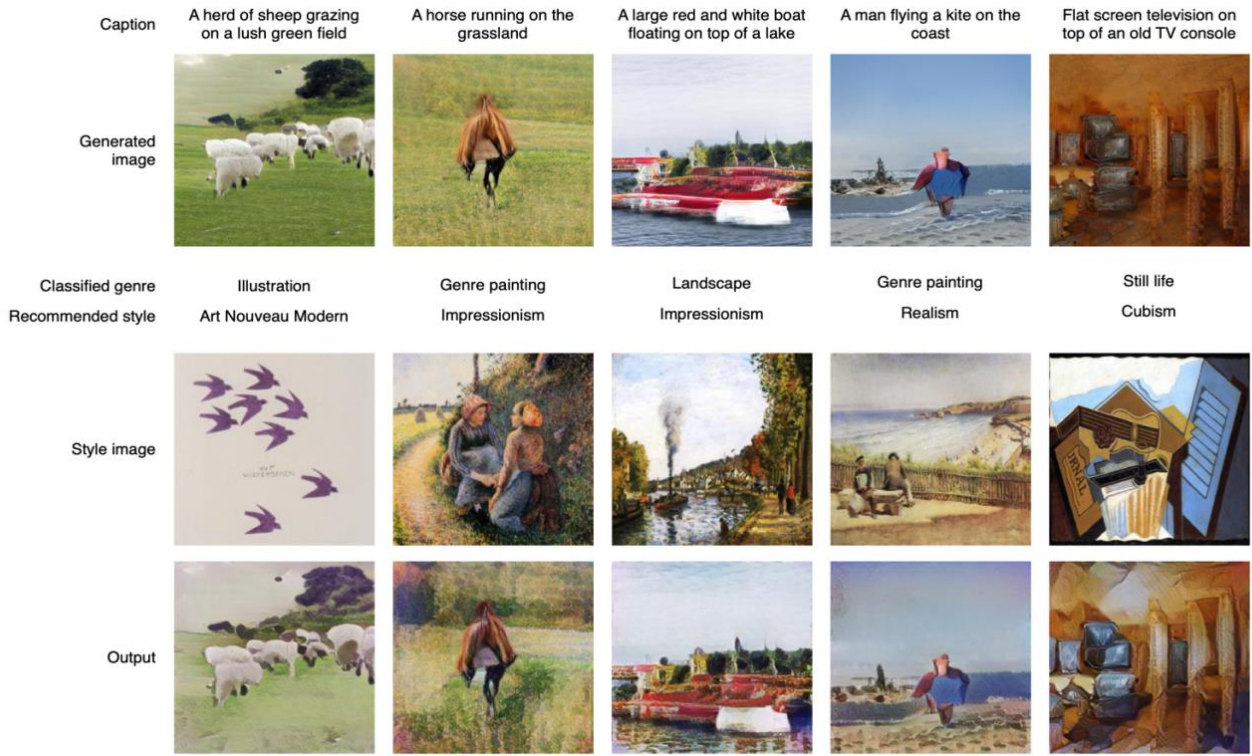

Figure 6. The overall result of our project.

For the style transfer and prediction networks, we tested our trained models across 45502 paintings for 27 styles in the *WikiArt* dataset and the pre-trained model based on *Describable Textures Dataset* (DTD) (Cimpoi et al. 2014), *Kaggle Painter By Numbers* (PBN) ("Painter by Numbers" n.d.) provided by (Ghiasi et al. 2017). Table 2 shows the raining result. When comparing the two models, we found that although the model pre-trained on DTD and PBN receives a style loss that is much lower than our model on observed styles (2.08e4 vs 7.48e5), the content loss is similar (8.92e4 vs 6.74e4). However, when testing on unobserved styles, both models achieve similar style loss (9.95e5 vs 1.07e6), and our model achieves a lower content loss (7.43e4 vs 6.48e4).

Finally, we combined the image generator, the classification network, and the style-transferer. Some of the results obtained are displayed in Figure 6.

| Dataset | Loss | Model A | Model B |
| --- | --- | --- | --- |
| Observed | Style loss | 7.48e5 | 2.08e4 |
|  | Content Loss | 6.74e4 | 8.92e4 |
| Unobserved | Style Loss | 1.07e6 | 9.95e5 |
|  | Content Loss | 6.48e4 | 7.54e4 |

Table 2: Performance of Style Transfer Networks
Model A: Model trained on WikiArt dataset
Model B: Model rained on DTD and PBN dataset

## 5 Graphical User Interface

A graphical user interface (shown in Figure 7) allows users to implement all three steps in an integrated end-to-end environment. After inputting the image description, the algorithm automatically generates a corresponding image and randomly select paintings from suitable artistic styles. If users are not satisfied with the selected artworks, they can repeat the selection process. The overall workflow of the program is presented in Figure 8.

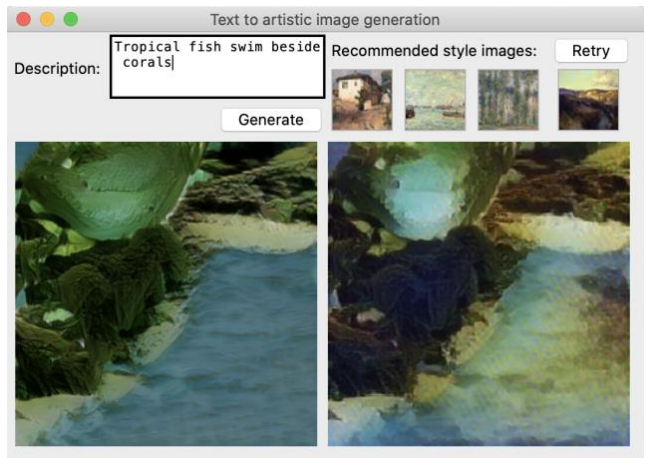

Figure 7. A graphical user interface written in Python

# 6 Conclusion and Future Works

In this paper, we combine various frameworks to generate painting-like artistic images from text descriptions. In general, we obtain acceptable results for multiple combinations of text inputs and desired styles. However, there are still many areas of our solution that can be improved. In particular, we plan to add a speech recognition module to make it possible for people with disabilities in hands to specify their inputs via voice instead of typing, which might be difficult. Furthermore, we would like to compare the various parts of our solution with that of other similar projects. Through this process, we hope to enhance the algorithms that were used so as to generate better quality.

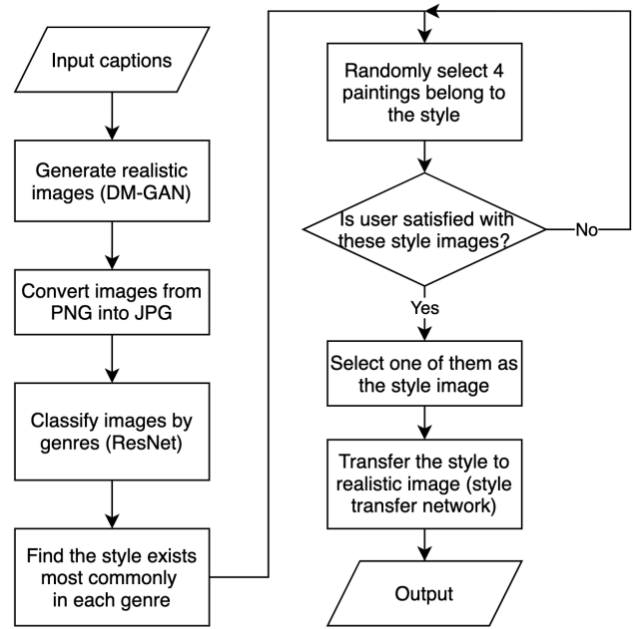

Figure 8. The overall steps of the project.